# Breast Cancer Recurrence Risk Prediction Based on MIL

**Jinqiu Chen[1][†][*], Huyan Xu[2][†][*]**

[1]College of Information Science and Technology, Beijing University of Chemical Technology, Beijing, China.

[2]School of Computer Science and Informatics, University of Liverpool, Liverpool, U.K.

**\*Correspondence:**
Jinqiu Chen, Huyan Xu
cjq3172629360@163.com; sghxu32@liverpool.ac.uk;
[†]These authors contributed equally to this work

## Abstract

Predicting breast cancer recurrence risk is a critical clinical challenge. This study investigates the potential of computational pathology to stratify patients using deep learning on routine Hematoxylin and Eosin (H&E) stained whole-slide images (WSIs). We developed and compared three Multiple Instance Learning (MIL) frameworks—CLAM-SB, ABMIL, and ConvNeXt-MIL-XGBoost—on an in-house dataset of 210 patient cases. The models were trained to predict 5-year recurrence risk, categorized into three tiers (low, medium, high), with ground truth labels established by the 21-gene Recurrence Score. Features were extracted using the UNI and CONCH pre-trained models. In a 5-fold cross-validation, the modified CLAM-SB model demonstrated the strongest performance, achieving a mean Area Under the Curve (AUC) of 0.836 and a classification accuracy of 76.2%. Our findings demonstrate the feasibility of using deep learning on standard histology slides for automated, genomics-correlated risk stratification, highlighting a promising pathway toward rapid and cost-effective clinical decision support.

## 1    Introduction

Breast cancer remains one of the most commonly diagnosed malignancies among women worldwide and continues to pose a substantial public health burden. According to global surveillance data released by the International Agency for Research on Cancer (IARC), it ranks first in incidence among female cancers, with a persistently rising trend observed in recent years (1). Despite steady advances in screening, surgery, systemic therapy, and supportive care, the clinical challenge of recurrence has not been resolved. For cases treated with early surgical resection, Antonios Valachis and colleagues reported 5-year and 10-year recurrence rates of 6.1% and 12.7%, respectively (2). Complementary population-level estimates from the SEER program indicate that the overall recurrence rate in the United States is approximately 30% (3). Recurrence and metastatic progression remain pivotal determinants of long-term survival and prognosis.

Accurate estimation of recurrence risk is therefore clinically consequential for both physicians and patients. On one hand, robust stratification provides an evidence base for individualized therapy: patients identified as high risk may benefit from treatment intensification—such as escalated chemotherapy or prolonged endocrine therapy—to reduce the likelihood of recurrence; conversely, patients categorized as low risk may safely undergo de-escalation, avoiding unnecessary toxicity and financial burden (4). On the other hand, precise risk prediction informs longitudinal management. When risk can be quantified with greater confidence, high-risk patients can be prioritized for closer surveillance, more frequent diagnostic assessments, and targeted rehabilitative guidance, with the potential to improve quality of life and extend survival.

Histopathological examination of tissue sections remains the diagnostic gold standard in breast oncology. Nevertheless, conventional pathology has notable limitations in the context of risk assessment. Decisions are largely based on morphological patterns and cellular features observed on stained slides, which are inherently sensitive to the observer's experience and thus subject to inter- and intra-observer variability, particularly in complex cases (5). The interpretive process is also labor-intensive, requiring exhaustive review of large, information-dense images. Moreover, subtle morphologic cues and weak signals related to underlying molecular alterations may be difficult to detect consistently by eye, constraining the precision of recurrence risk estimation.

Rapid progress in artificial intelligence has catalyzed new approaches to medical image analysis. Deep learning, in particular, has demonstrated strong performance in tasks such as detection, classification, and segmentation across radiology and pathology. In whole-slide histopathology, data-driven models can learn hierarchical representations directly from large image corpora and support reproducible, quantitative characterization of tissue and cellular architecture, and therefore be able to predict recurrence risk from these learned features.

There is a notable challenging in predicting recurrence risk of breast cancer using deep learning. Private quality datasets were used to train deep learning models in previous researches, which contains large mount of WSIs and annotations. However, these datasets are costly and time-consuming to acquire, and exhibit limited universality. As a result, they remain inaccessible to institutions and clinics with constrained resources. Pretrained models have demonstrated strong adaptability to perform a wide range of tasks. Nevertheless, the application of specialized pretrained models in the field of computational pathology has not been thoroughly explored and applied in prior studies.

To address the above challenge, we leverage pretrained models to reduce data dependency. First, we adopt UNI and CONCH to extract features of WSIs from a small dataset with merely 210 images. Then we explore three models with different architectures to predict breast cancer recurrence risk based on features extracted by pretrained models. As for results, all models achieve accuracies above 70%, manifesting the potential of using pretrained models to reduce the need for data. The main contributions of this work are as follows:

- We utilize UNI and CONCH to extract features from WSIs.

- We propose the CLAM-SB model, ABMIL model and ConvNeXt-MIL-XGBoost model, achieving accuracies of 76.2%, 70.9%, 73.5% respectively.

## 2    Related Works

In recent years, rapid advances in whole-slide digital scanning have transformed conventional glass slides into high-fidelity whole-slide images (WSIs). Modern scanners digitize the entirety of stained tissue sections and store them as multi-level files proportional to image size. This capability allows pathologists to conduct diagnostic review directly on screen, substantially improving efficiency and quality of assessment. More importantly, WSI technology has enabled computational approaches, including deep learning, to be systematically integrated into pathology workflows. WSIs are typically archived in a multi-resolution pyramid (for example, 20×, 10×, 5 ×), which supports seamless navigation from global tissue context down to cellular detail, thereby offering new perspectives for diagnostic decision-making and research (6).

Jiang et al. explored machine learning on multimodal and multi-omics data to characterize recurrence risk in breast cancer (7). The team developed a multimodal model termed TMPIC that stratifies recurrence risk into high- and low-risk groups. Validation on a held-out test set indicated strong predictive performance, facilitating the identification of patients at elevated risk who may benefit from intensified management.

Goyal et al. proposed OncoDHNet, a transformer-based deep learning model that integrates morphologic patterns in pathology images with clinicopathologic variables to predict recurrence risk (8). Their evaluation drew on internal cohorts from Dartmouth Health and the University of Chicago, comprising 990 WSIs and clinical-pathologic data from 981 patients. The study underscores the promise of hybrid models that fuse image-derived representations with structured clinical information.

Whitney et al. analyzed H&E-stained images from 178 early-stage, estrogen receptor–positive breast cancer patients to examine nuclear morphology at scale (9). They extracted 216 nuclear morphometric features and assessed four classifiers—random forests, neural networks, SVM, and linear discriminant analysis—combined with four feature selection strategies (Ranksum, PCA-VIP, mRMR MID, mRMR MIQ). Reported test-set accuracies ranged from 75% to 86%, indicating that carefully engineered features coupled with classical machine learning can yield competitive performance in recurrence risk prediction.

Despite these advances, notable gaps remain. Public WSI datasets specific to breast cancer recurrence are still modest in scale, constraining model training and limiting generalization across institutions and scanners. Moreover, many deep models operate as complex "black boxes", which complicates clinical interpretability and slows translation into routine practice (10). Addressing data scarcity, domain shift, and explainability will be essential to bridge methodological progress and clinical impact.

## 3 Method

### 3.1 Datasets Introduction

The overall workflow of this study is illustrated in Figure 1. It encompasses several key stages: dataset acquisition and preparation, model development and training, performance evaluation, and finally, visualization and interpretability analysis.

The dataset for this research was sourced from the Department of Pathology at the China-Japan Friendship Hospital. It consists of a cohort of 210 breast cancer patients, for whom Hematoxylin and Eosin (H&E) stained whole slide images (WSIs) were available. The selection of these cases was performed under the guidance of experienced pathologists.

The ground truth for recurrence risk assessment was established using the 21-gene Recurrence Score (e.g., Oncotype DX). This genomic assay quantifies the expression levels of 21 specific genes within the tumor tissue to predict the likelihood of recurrence for patients with early-stage, hormone receptor-positive (HR+), HER2-negative breast cancer (11). For the purpose of this study, the prediction endpoint was defined as the risk of disease recurrence within a five-year period following the initial diagnosis. This score is a clinically accepted standard for guiding adjuvant chemotherapy decisions.

Based on this endpoint, the recurrence risk scores were categorized into three distinct labels for our three-tier classification model: 0 for Low Risk, 1 for Medium Risk, and 2 for High Risk. All training and validation procedures in this study were conducted exclusively on this institutional dataset. The WSIs were digitized using a slide scanner from Shengqiang Technology Co., Ltd. (Shenzhen, China), which produces images in a proprietary .spdc format designed for high-resolution digital pathology.

### 3.2 WSI Preprocessing

The WSIs obtained for this study were initially in a proprietary .sdpc format. As this format is not compatible with standard open-source Python libraries for digital pathology, an initial conversion step was necessary. Using the vendor-provided software, all .sdpc files were converted to the more widely supported .svs format to facilitate subsequent processing.

Given the gigapixel resolution of WSIs, which makes direct processing computationally infeasible, a comprehensive preprocessing pipeline was implemented (summarized in Figure 2). The primary goal of this pipeline was to accurately segment the tissue regions from the background of the glass slide.

The segmentation process began by selecting an optimal low-magnification level from the WSI's pyramidal structure, automatically identified using the get_best_level_for_downsample function from the OpenSlide library. On this downsampled image, an adaptive Gaussian blur was applied to mitigate noise arising from staining inconsistencies. Subsequently, the image was transformed from the RGB to the HSV (Hue, Saturation, Value) color space. The Saturation (S) channel was specifically chosen for segmentation, as it provides high contrast between the hematoxylin (blue) and eosin (pink) stains.

To binarize the image and create a tissue mask, we employed a modified Otsu's thresholding algorithm, enhanced with morphological gradients. This was followed by a series of morphological filtering operations to remove staining artifacts and filter out small, detached tissue fragments, ensuring that only contiguous and significant tissue areas were retained for analysis.

After generating the final tissue mask, a sliding window approach was used to extract patches from the original high-resolution WSI. Non-overlapping patches of 256×256 pixels were extracted from all regions covered by the tissue mask. The coordinates of these patches were systematically stored in HDF5 files for efficient retrieval during model training. To ensure the quality of our data pipeline, the segmentation masks were visualized for manual inspection, allowing for verification of the segmentation accuracy before proceeding to feature extraction, as exemplified in Figure 3.

### 3.3 Ground Truth Labeling

For this study, several potential sources for ground truth labels were considered, including: (1) the raw 21-gene Recurrence Score, (2) the three-tiered recurrence risk (Low, Medium, High) determined by pathologists, which integrates the Recurrence Score with other clinicopathological factors, and (3) the binary outcome of recurrence or death based on two-year follow-up data.

After careful consideration and consultation with pathologists, we selected the pathologists' integrated three-tiered risk assessment as the definitive ground truth for training our classification model. This decision was based on several key factors. First, relying solely on the 21-gene Recurrence Score as the label would not fully reflect the real-world diagnostic process, where clinicians make a comprehensive judgment by considering the score in conjunction with other crucial pathological features (as detailed in Table 1).

Second, using the binary follow-up status (recurrence/death) was deemed unsuitable due to significant data incompleteness. A substantial portion of patients had missing follow-up records. Furthermore, among the available records, only a small number of patients (just over ten) were documented as having experienced recurrence or death. This low count was considered by pathologists to be an underrepresentation of the true event rate, posing a high risk of small-sample bias and model overfitting. Such a label would fail to capture the prognostic patterns associated with the high-risk patient group accurately.

In contrast, the three-tiered risk classification provided a more complete and balanced dataset for a supervised learning task. While the Medium-risk category was relatively small (21 out of 210 cases), the Low- and High-risk classes were more evenly distributed. This distribution was deemed more suitable for training a model capable of learning discriminative features from the WSI data.

### 3.4 Whole Slide Image Feature Extraction

Deep learning-based prediction of cancer recurrence from WSIs is a challenging task, primarily due to the computational bottleneck associated with feature extraction from gigapixel-scale images. The quality of the features extracted at this stage is critical, as it directly impacts the performance of any subsequent classification model.

To address this challenge, we employed the TRIDENT toolbox, an open-source framework for digital pathology developed at Harvard University. TRIDENT provides a comprehensive suite of tools for processing WSIs, including tissue segmentation, patch-level feature extraction using various pre-trained models, and slide-level feature aggregation. For this study, we selected two foundation models available within TRIDENT to serve as our feature extractors: UNI and CONCH.

UNI (Universal Network Initiative) is a large-scale, pre-trained vision encoder designed explicitly for computational pathology by the Mahmood Lab. It is built upon a Vision Transformer (ViT-L/16) architecture and was trained using a self-supervised learning method (DINOv2) on an extensive dataset of 100 million image patches and 100,000 WSIs, covering a diverse range of tissues including cancerous, inflammatory, and normal samples (12).

CONCH (Contrastive Learning from Captions for Histopathology) is a vision-language foundation model also developed for computational pathology. By learning from paired image and text (pathology report) data, CONCH is designed to extract rich, context-aware features even in scenarios with limited annotations, emulating the diagnostic workflow of a pathologist (13).

Rationale for Model Selection

The selection of both UNI and CONCH was a deliberate methodological choice based on the hypothesis that their distinct pre-training paradigms would capture a complementary set of histopathological features. UNI, trained via self-supervision on a massive corpus of images, excels at learning general-purpose, data-driven visual patterns directly from pixel data. In contrast, CONCH, as a vision-language model, is trained to associate morphological features with descriptive pathological text, potentially enabling it to learn more semantically rich and context-aware representations. While other powerful Vision Transformer-based models such as CTransPath have shown excellent performance, our strategy was to leverage the potential synergy between a pure visual encoder (UNI) and a vision-language model (CONCH) to create a more comprehensive feature set for the downstream classification task. Furthermore, at the time this research was initiated, both models represented the state-of-the-art and were readily accessible within the TRIDENT framework, facilitating their direct integration into our pipeline.

The decision to use these powerful pre-trained models was motivated by the limited size of our in-house dataset (210 WSIs). By leveraging transfer learning, we can harness the knowledge encoded within these models from their training on massive datasets, which is essential for developing a robust model on a smaller cohort.

The feature extraction pipeline was implemented as follows: leveraging the tissue masks and patch coordinates generated in the preprocessing stage (Section 3.2), each 256x256 pixel patch was resized to match the input dimensions required by the UNI and CONCH models. The pre-trained models were then used to process each patch sequentially, generating a feature vector for each one. These patch-level feature vectors were then systematically saved into HDF5 files, preparing them for the subsequent multi-instance learning stage.

## 3.5   Model Development and Optimization

### 3.5.1 The CLAM-SB Model Architecture

Our first model is based on CLAM (Clustering-constrained Attention Multiple instance learning), an attention-based Multiple Instance Learning (MIL) framework designed to automatically identify diagnostically relevant sub-regions for accurate slide-level classification (14).

The MIL paradigm is particularly well-suited for computational pathology. In this context, a WSI is treated as a "bag," and its constituent patches are treated as "instances." The model's objective is to predict a single label for the entire bag. The CLAM model achieves this by learning the importance weights of different instances via an attention mechanism, which then guides the aggregation of instance features for the final classification. The architecture comprises three main components: a feature encoder (the pre-trained UNI/CONCH models), an attention aggregator, and a classifier.

A core component of CLAM is its gated attention mechanism, which is defined as:

$$A = \sigma(W_b h + b_b) \odot \tanh(W_a h + b_a) \tag{1}$$

where $W_a$ and $W_b$ are learnable weight matrices, and $\odot$ denotes element-wise multiplication. This structure uses a sigmoid function to generate a spatial attention gate, which modulates the feature response activated by a tanh function. This gated approach allows for a more nuanced capture of subtle pathological features compared to standard attention mechanisms. The resulting attention scores are then used to perform a weighted aggregation of all instance features into a single slide-level feature vector, which encapsulates the global semantic information of the WSI.

Modifications to the CLAM Architecture:

For our study, we introduced several significant modifications to the baseline CLAM model to enhance its feature learning capacity on our complex WSI dataset.

1). Deeper Network Structure: We deepened the classifier by adding an extra fully connected layer, introducing a more profound non-linear transformation to better learn complex interactions between features.

2). GELU Activation: The standard ReLU activation was replaced with the Gaussian Error Linear Unit (GELU) (15). GELU is defined as:

$$GELU(x) = x \cdot \Phi(x) \tag{2}$$

where $\Phi(x)$ is the cumulative distribution function of the standard normal distribution. A common approximation is used in practice:

$$GELU \approx \frac{1}{2} x \left[ 1 + \tanh \left( \sqrt{\frac{2}{\pi}} (x + 0.044715 x^3) \right) \right] \tag{3}$$

Unlike ReLU's hard truncation, GELU's smooth gradient around x≈0 can help alleviate the vanishing gradient problem and facilitate more stable training.

3). Enhanced Attention Network: The hidden dimension of the attention network was increased from 256 to 384, bolstering the model's capacity to capture high-dimensional pathological feature representations.

4). Advanced Regularization: We implemented a more aggressive regularization strategy by increasing the Dropout rate within the classifier. Furthermore, Dropout was applied independently to the feature encoding, attention, and classification modules to prevent feature co-adaptation. This was particularly important for mitigating overfitting on the underrepresented medium-risk class.

5). Multi-Layer Classifier: The main classifier was expanded from a single layer to a two-layer network. This included an additional hidden layer with a non-linear activation and Dropout, enhancing its ability to learn a more complex decision boundary. We also introduced dimensionality reduction (from 512 to 256) in the intermediate layer, compelling the model to learn a more discriminative and compact feature representation. The modified architecture is illustrated in Figure 4.

Training Strategy for CLAM-SB:

The hyperparameters used for training, such as learning rate and batch size, are detailed in Table 2. The overall training workflow is depicted in Figure 5.

A primary challenge was the severe class imbalance, with the medium-risk class constituting only 10% of the dataset (21 of 210 cases). To address this, we employed a multi-faceted strategy. The model was trained using the Adam optimizer, and we incorporated the Focal Loss function, which is specifically designed for scenarios with extreme class imbalance (16). The Focal Loss is defined as:

$$FL(pt) = -\alpha t(1 - pt)\gamma \log(pt) \qquad (4)$$

We applied a high-class weight of $\alpha t = 3.0$ to the medium-risk category. The focusing parameter $\gamma$ dynamically down-weights the loss for well-classified examples, thereby forcing the model to concentrate on hard, low-confidence samples (e.g., prediction probability $pt < 0.3$). This strategy significantly enhances the model's sensitivity to the minority class.

Furthermore, to improve model calibration and prevent overfitting to the few medium-risk samples, we utilized label smoothing (17). The one-hot encoded ground truth labels were converted to a soft distribution using a smoothing factor of $\epsilon = 0.1$, defined as

$$y' = (1 - \epsilon)y + \epsilon \cdot \frac{1}{K} \qquad (5)$$

where $K$ is the number of classes ($K = 3$ in our case).

## 3.5.2 The ABMIL Model: Architecture and Training

In addition to the CLAM-based framework, we developed an independent classification model inspired by the Attention-based Multiple Instance Learning (ABMIL) architecture (18). This model follows a canonical encoder-aggregator structure, where each patient's WSI represents a "bag" and the extracted patches serve as "instances" within that bag.

Model Architecture:

The core of our custom ABMIL model is a gated multi-head attention network. This network consists of eight parallel attention heads, allowing the model to capture a diverse range of feature patterns simultaneously. Within each head, the high-dimensional instance features (1024-dim) are first projected into a lower-dimensional attention space (256-dim) and then passed through a Tanh activation function to generate a base attention vector.

To enhance the model's ability to focus on critical regions, we incorporated a gating mechanism into each attention head. This is implemented as a sigmoid-activated layer that produces a gating weight between [0, 1]. This gate is then multiplied element-wise with the base attention vector. This design emulates the excitation-inhibition mechanism of biological neurons, enabling the model to dynamically suppress interference from noisy or irrelevant regions while amplifying the signal from key pathological features.

A key modification in our design is the method for calculating attention scores. Instead of using a conventional dot-product between query and key vectors, we employ a linear layer to directly compute attention scores for each of the three risk classes (Low, Medium, High). These class-specific scores are then normalized across all instances in the bag using a softmax function. This approach allows the model to learn distinct spatial attention maps corresponding to the morphological patterns of each risk level. We hypothesize that this direct mapping is more flexible and robust for smaller medical imaging datasets, as the linear layer's weights can be learned to capture complex data-driven relationships.

The aggregation process utilizes a three-branch strategy. The instance features, weighted by their class-specific attention scores, are aggregated independently for each of the three risk classes. This results in three distinct bag-level feature vectors (each 1024-dim), where each vector represents the salient pathological evidence for a specific risk category.

Finally, these three aggregated vectors are passed to a feature enhancement network for final classification. This network employs a bottleneck structure ($1024 \rightarrow 512$ dimensions) to reduce redundancy and learn a more discriminative representation. The output from this bottleneck is then fed into a fully connected layer that maps the features to the final probability distribution over the three risk classes. The complete architecture of our modified ABMIL model is depicted in Figure 6.

Model Training and Validation:

The model was trained end-to-end using the features extracted by the CONCH encoder. To address the class imbalance, we utilized a weighted Cross-Entropy loss function, which explicitly increases the loss contribution from the underrepresented medium-risk class. The model parameters shown in Table 3 were optimized using the Adam algorithm.

During training, the gated multi-head attention module learns to identify significant pathological patterns via backpropagation. The gating layers dynamically adjust their selectivity; we observed that during the initial 50% of training epochs, the gates tend to remain more open to facilitate a broad exploration of the feature space, while in later stages, they become more selective to focus on the most discriminative regions.

### 3.5.3 The ConvNeXt-MIL-XGBoost Model

To address the challenges of recurrence risk prediction based on whole-slide images (WSIs), we propose a three-stage architecture that decouples visual representation learning, instance-level aggregation, and final decision making. This modular design enables more effective handling of high-resolution image data, class imbalance, and interpretability — three central issues in medical image analysis.

Figure 7 shows the overview of the proposed model (1) Patch-level feature extraction using ConvNeXt-Base (2) Slide-level representation learning via attention-based Multiple Instance Learning (MIL). (3) Final classification using the gradient-boosted tree model XGBoost.

The overall pipeline draws on principles of hierarchical modeling and ensemble decision making, which have been demonstrated to be effective in various vision tasks, including pathology (19).

Stage One: Patch Feature Extraction with ConvNeXt:

Whole-slide images are subdivided into 256×256 patches using a sliding window strategy, with background areas excluded by tissue segmentation. For patch-level representation learning, we utilize the ConvNeXt-Base model, a convolutional neural network incorporating modern design patterns derived from vision transformers, such as large kernel sizes and normalization-based scaling (20).

ConvNeXt has demonstrated state-of-the-art performance in natural image classification and is well-suited for medical tasks due to its balance of locality and global receptive field modeling. Each patch is encoded into a 1024-dimensional feature vector. Compared to earlier CNNs such as ResNet, ConvNeXt offers improved capacity for capturing multi-scale tissue textures and structural context, which is critical for identifying histological patterns.

Initial experiments using ResNet-18 as the encoder in our baseline pipeline showed limitations in both expressiveness and generalization, especially in handling the heterogeneity of tumor regions. ConvNeXt was therefore selected as the backbone to enhance visual feature learning.

Stage Two: Slide-Level Aggregation via Attention-based MIL:

Given that annotations are only available at the slide level, we employ an attention-based Multiple Instance Learning (MIL) strategy to model the relationship between patch-level features and slide-level labels. Each slide is treated as a bag of instances (patches), and attention weights are computed to quantify the relevance of each patch in determining the overall diagnosis. Our design of the attention module is inspired by recent domain-specific work in colorectal cancer histopathology, where attention-based patch aggregation has been successfully used to identify tumor budding patterns (21).

The MIL framework computes a weighted average of patch features, forming a fixed-length slide embedding. Let $\{x_1, x_2, \ldots, x_n\}$ denote the patch features and $a_i$ their corresponding learned weights. The aggregated vector is given by $S = \sum a_i x_i$. This embedding not only serves as a compact representation of the WSI but also supports interpretability by identifying key regions via attention maps.

To address class imbalance, we incorporate focal loss during training, which down-weights the loss from well-classified examples and emphasizes learning from hard samples (16). This is particularly helpful in medical settings where minority classes (e.g., high-risk cases) are underrepresented.

Stage Three: Final Prediction via XGBoost:

For classification, we utilize XGBoost, a gradient-boosted decision tree model that excels at modeling structured features and is robust to sample imbalance. The input to XGBoost consists of 1024-dimensional slide-level embeddings from the MIL stage, optionally concatenated with 23 handcrafted statistical features derived from patch-wise attention distributions.

XGBoost's learning objective combines loss minimization with regularization to avoid overfitting. Its inherent ability to select salient features and model nonlinear interactions is beneficial in our setting, where the representations may contain both redundant and highly discriminative dimensions.

Formally, the model minimizes:

$$L = \sum_i l(y_i, \hat{y}_i) + \sum_k \Omega(f_k) \text{ , where } \Omega(f) = \gamma T + \frac{1}{2}\lambda \|w\|^2 \tag{6}$$

where $T$ is the number of leaves and $w$ the leaf weights. This formulation encourages model sparsity and prevents overfitting.

The application of XGBoost is inspired by recent work in tumor detection, where convolutional and gradient-boosted architectures were effectively combined for enhanced classification performance in limited-data settings (19).

Summary and Inter-Stage Synergy:

Each module in the pipeline contributes a distinct function: ConvNeXt captures rich local and contextual features; MIL dynamically selects informative regions under weak supervision; and XGBoost performs interpretable, regularized classification. This design ensures resilience against patch redundancy, class imbalance, and model overfitting, while also providing interpretability via attention and feature importance analysis.

The pipeline achieves a classification accuracy of 75.0% with an F1 score of 0.680, and outperforms the baseline in both stability and sensitivity. The modularity of the system also facilitates future extensions such as backbone substitution or the integration of clinical metadata.

## 4 Results

### 4.1 Results of CLAM-SB and ABMIL

A 5-fold cross-validation protocol was employed for robust performance evaluation, a necessary approach given the limited cohort size. This mitigated partitioning bias and provided a reliable estimate of model generalization. Both models demonstrated successful training dynamics, achieving stable convergence as indicated by their respective learning curves(Figure 8 ,Figure 9 ,Figure 10).

Quantitative Performance Analysis:

The classification performance of the models was evaluated on the validation set of each fold. The CLAM-SB model achieved a promising level of performance, reaching an accuracy of over 80% on several folds, with an average accuracy exceeding 75% across all five folds (detailed metrics in Table 4).

An analysis of the best-performing fold provides a more granular view of the model's capabilities. In this fold, the model correctly classified all low-risk cases (12 out of 12) and a majority of high-risk cases (4 out of 7). This demonstrates the model's strong ability to distinguish the morphological features separating the low- and high-risk categories in unseen WSIs. However, performance on the medium-risk class was notably weaker, with only one of the two cases being correctly identified. This indicates that despite the implementation of class-weighting and Focal Loss, the model's ability to recognize the underrepresented medium-risk class remains a challenge.

The ABMIL model achieved a peak accuracy of 77% on its best-performing fold, with an average accuracy of approximately 70% across all folds (detailed metrics in Table 5). The per-class performance on its best fold revealed a similar pattern: it achieved perfect accuracy for the low-risk class (100%) and strong accuracy for the high-risk class (71%). However, its primary limitation was a complete failure to identify any medium-risk cases (0% accuracy). This suggests that, compared to the CLAM architecture, our ABMIL model was more susceptible to the severe class imbalance, even with countermeasures in place. The poor performance on the medium-risk class for both models is likely attributable to its small sample size, constituting only 10% of the entire dataset, which may have been insufficient for the models to learn a robust feature representation.

Interpretability Analysis via Attention Heatmaps:

To enhance the transparency of our models and provide insight into their decision-making processes, we generated attention heatmaps to visualize the regions within each WSI that most influenced the final prediction.

For the CLAM model, the interpretability analysis was conducted by leveraging its inherent attention mechanism. After feeding the patch-level feature vectors into the trained model, the attention scores for each patch were extracted. These scores, which represent the diagnostic importance of each region, were then mapped back to their original coordinates on the WSI. The resulting sparse attention map was smoothed using a Gaussian filter and visualized using a "Jet" colormap. This heatmap was then overlaid on the original H&E slide to produce an intuitive visualization of the model's focus (Figure 11), with detailed high-magnification views revealing patch-level attention scores (Figure 12).

For the ABMIL model, a similar visualization pipeline was implemented. During inference on a WSI, the raw attention tensors were retrieved from the model's multi-head attention module. These patch-level attention weights were then upsampled to the original WSI resolution using bilinear interpolation to create a continuous heatmap. This map was subsequently overlaid onto the H&E image, highlighting the areas the model deemed most salient for its prediction (Figure 13), with corresponding zoomed-in views for detailed inspection (Figure 14).

These attention-based visualizations provide a crucial layer of explainability, potentially aiding pathologists in verifying the model's predictions and building trust in its diagnostic utility.

## 4.2    Results of ConvNeXt-MIL-XGBoost

### 4.2.1 Overall Model Performance

The proposed three-stage MIL-XGBoost framework achieved a classification accuracy of 73.5% on the independent test set, with a macro F1-score of 0.492, demonstrating reasonable classification capability given the complexity of whole-slide histopathology data and the limited sample size (19). The model maintained consistent performance across training and testing, indicating effective generalization.

Figure 15 presents the overall accuracy and macro F1-scores across training, validation, and test sets, highlighting the model's generalization capabilities.

### 4.2.2 Performance by Risk Category

Despite class imbalance, the model exhibited particularly strong performance in low-risk classification, achieving an F1-score of 0.844, with precision of 76.0% and recall of 95.0%, underscoring its ability to confidently identify benign cases. This is of particular importance in clinical screening scenarios, where accurate detection of low-risk patients can substantially reduce the diagnostic burden on pathologists.

While medium- and high-risk classifications exhibited varying levels of accuracy, these should be interpreted with caution due to the small number of annotated samples in these categories. Notably, the model correctly predicted a majority of high-risk cases (F1 = 0.632) with acceptable precision (75.0%), even under class imbalance and data scarcity.

Figure 16 illustrates the confusion matrix on the test set, indicating high precision and recall in the low- and high-risk categories.

### 4.2.3 Model Configuration and Feature Utilization

The final XGBoost classifier was trained on a concatenated feature vector composed of 23 enhanced features. These included MIL-derived logits and probabilities, attention statistics, patch count, and distribution-level descriptors—forming a comprehensive feature set that encapsulated both local and global patterns in the WSIs. The ensemble utilized 200 decision trees with a learning rate of 0.1 and maximum tree depth of 6.

Figure 17 shows the top-ranked features in the XGBoost classifier, demonstrating the contribution of MIL attention distribution and slide-level features.

### 4.2.4 Evaluation Protocol and Data Leakage Control

To ensure the reliability of experimental findings, the study employed strict separation of training, validation, and test sets. All test evaluations were conducted on fully unseen samples, and hyperparameter tuning was performed using only the training and validation splits. This design effectively eliminated data leakage risks and reinforces the credibility of the reported results.

### 4.2.5 Computational Efficiency

The pipeline demonstrated high computational efficiency during both training and inference phases. ConvNeXt enabled parallelized patch-level feature extraction, while the MIL attention mechanism reduced redundancy by focusing on diagnostically relevant regions. XGBoost delivered rapid inference with robust decision boundaries, contributing to the framework's practical viability in real-world clinical settings.

## 5    Discussion

In this study, we developed and compared three distinct Multiple Instance Learning (MIL) frameworks—CLAM-SB, a custom ABMIL, and ConvNeXt-MIL-XGBoost—for the challenging task of predicting three-tiered breast cancer recurrence risk from H&E stained WSIs. Our principal finding is that the computational analysis of histopathological morphology can effectively stratify patients according to a genomically-defined risk profile. Our modified CLAM-SB model emerged as the superior architecture with a mean AUC of 0.836. Concurrently, the 73.5% accuracy achieved by the ConvNeXt-MIL-XGBoost model also highlights its robustness, particularly given the constraints of a small and imbalanced dataset. A key finding across our models was their strong capacity to differentiate between low- and high-risk categories, suggesting the presence of distinct, learnable morphological phenotypes for these risk extremes.

However, it is crucial to contextualize these results within the broader literature, which has documented several persistent challenges in applying MIL to computational pathology. Model performance is often sensitive to domain shift, including variations in tissue preparation, staining protocols, and digital scanners across different institutions, which can significantly hinder generalizability (14). Furthermore, while adept at identifying widespread patterns, attention-based MIL models can struggle with prognostic tasks that depend on subtle, sparsely distributed, or morphologically ambiguous features, as the attention mechanism may fail to capture these nuanced signals effectively (22). Our own models' difficulty in robustly identifying the medium-risk class aligns with these reported challenges. This suggests that while powerful, MIL is not a panacea; its success is intrinsically linked to the distinctness of the underlying histological patterns and the balance of the training data.

Despite these general challenges, our study possesses several notable strengths. Its foundation on a real-world, single-institution clinical dataset provides a realistic benchmark for model performance. The use of the 21-gene Recurrence Score as the ground truth offers a more objective and biologically grounded endpoint than subjective annotations or incomplete follow-up data. Furthermore, our work provides a valuable head-to-head comparison of different advanced MIL architectures. The focus on a clinically nuanced three-tiered classification, while difficult, also represents a step toward more granular patient stratification. Finally, the integration of attention heatmaps provides a crucial layer of interpretability, which is essential for building clinical trust.

These strengths notwithstanding, we acknowledge the study's limitations. The primary limitation is the severe class imbalance, which led to poor predictive performance for the medium-risk class. The small sample size (n=21) was likely insufficient for learning a robust feature representation for this intermediate profile. Another limitation is the single-center nature of our dataset, which, as discussed, restricts the generalizability of our findings. These limitations directly inform our future research directions. The most critical next step is to expand the dataset and validate the models on external, multi-center cohorts. Computationally, we plan to explore advanced techniques such as ordinal regression or generative data augmentation. Further enhancements could come from moving towards a multi-modal prognostic framework that integrates WSI features with other data streams like patient clinical data and key molecular markers.

Beyond these research-focused next steps, translating this work into a clinical-grade tool requires careful consideration of its integration into the existing pathological workflow. We envision our model operating as a decision support system within a hospital's digital pathology platform. Upon digitization of an H&E slide, the model could run in the background, providing the pathologist with a preliminary risk score and an attention heatmap integrated directly into their viewing software. This could serve as a "second-read" or a screening tool to flag cases with high-risk features for more detailed review or to help adjudicate borderline cases. However, significant practical challenges remain. These include obtaining regulatory approval (e.g., from the FDA or NMPA), ensuring seamless integration with diverse Laboratory Information Systems (LIS), and building the necessary computational infrastructure. The ultimate validation for such a tool would be its performance in large-scale, prospective clinical trials to demonstrate its real-world utility in improving patient outcomes.

## 6    Conclusion

In conclusion, this study demonstrates the feasibility of using attention-based MIL frameworks to predict breast cancer recurrence risk from routine histology slides. Our work highlights a promising pathway toward developing an AI-powered decision support tool that could complement existing genomic assays, potentially offering a rapid and cost-effective method for risk stratification. Future efforts focused on validating these models on larger, more diverse datasets and integrating multi-modal data will be critical for enhancing predictive accuracy and ultimately realizing their clinical utility.

**Figure and Table Legends**

Figure 1. Research Workflow of the Thesis

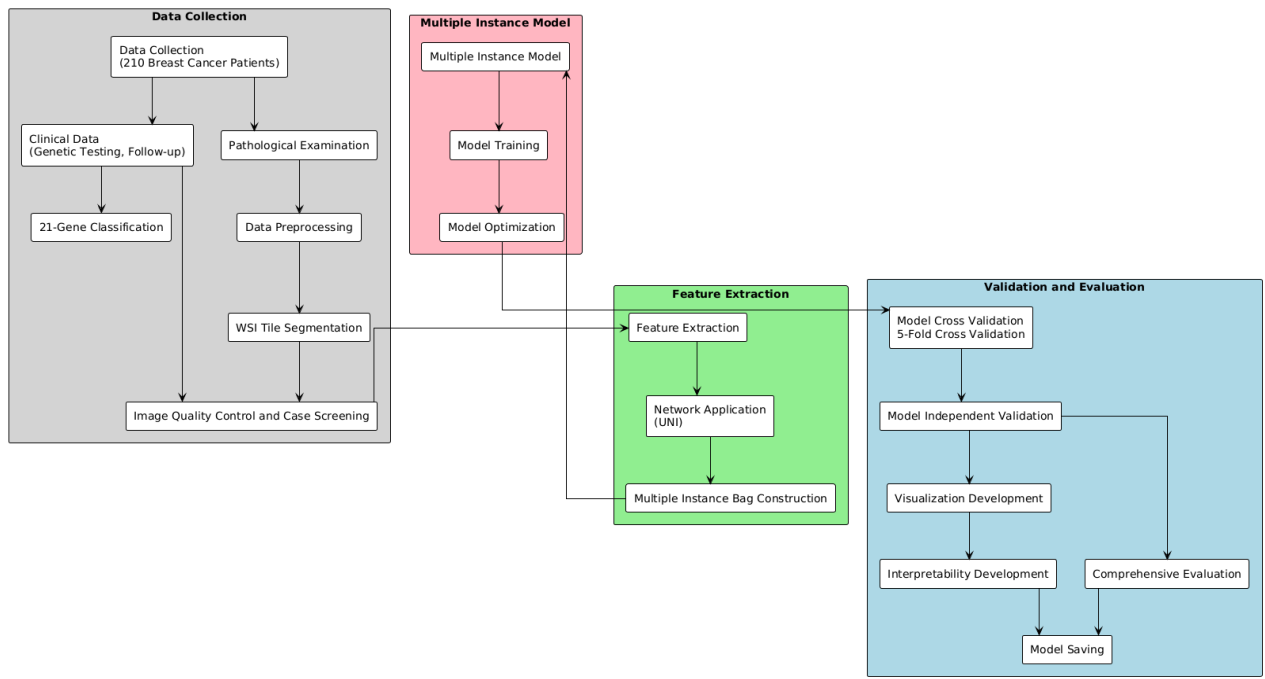

Figure 2. Preprocessing Workflow

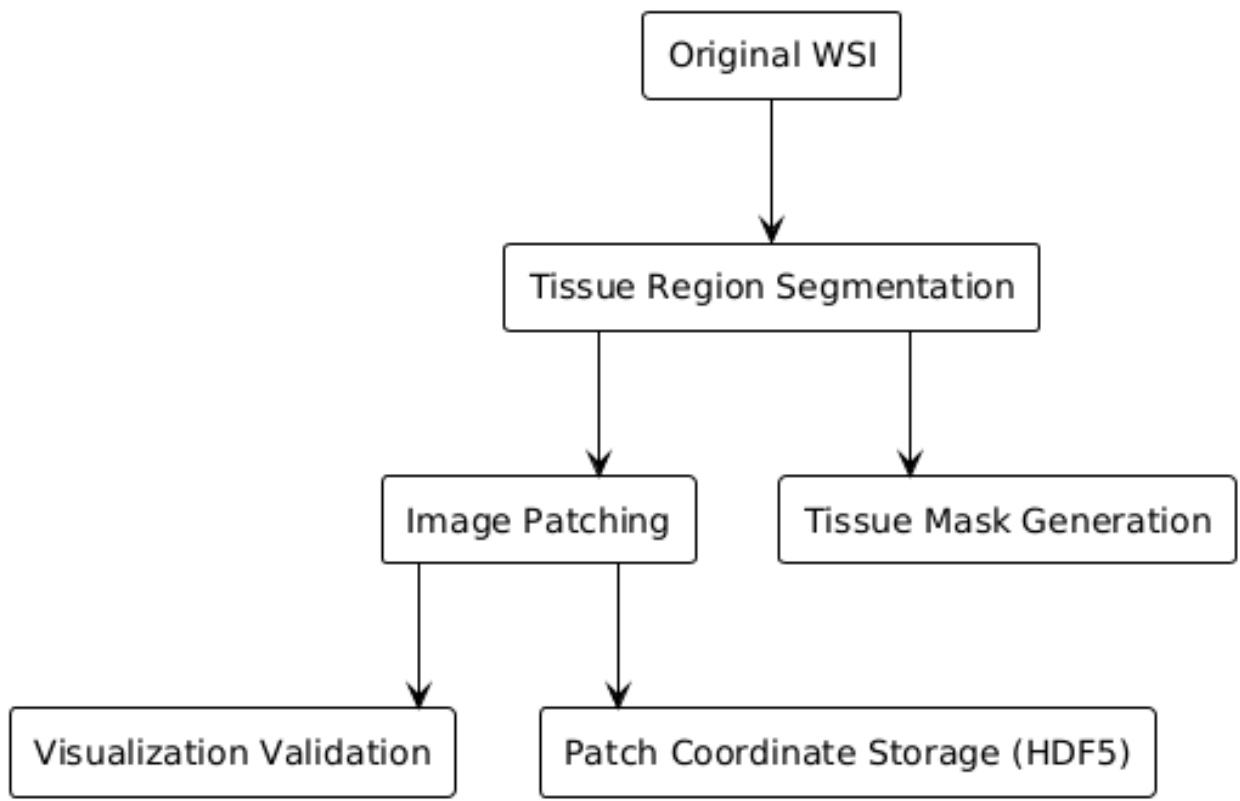

Figure 3. Visualization of WSI Segmentation

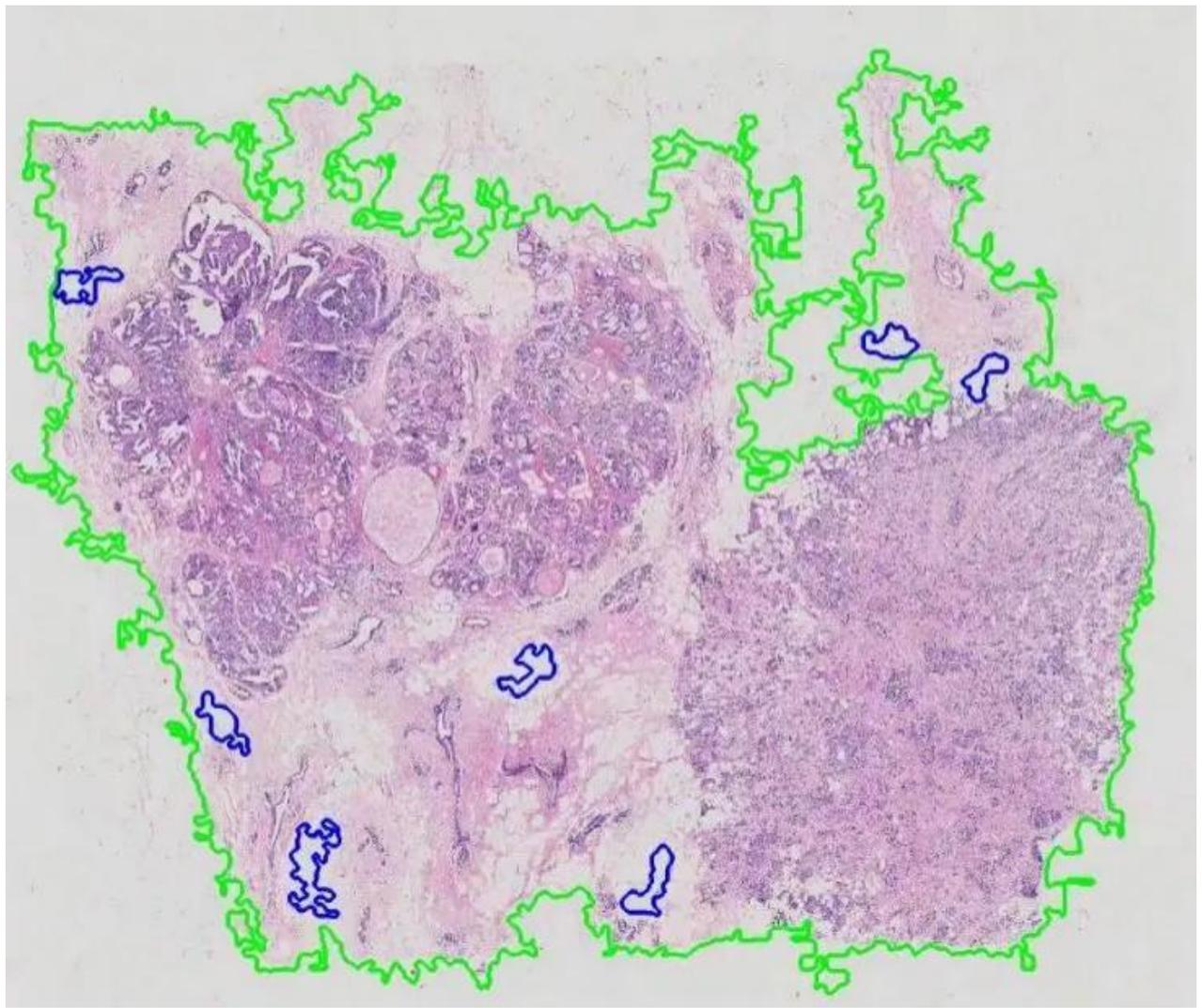

Figure 4. Fine-Tuned CLAM-SB Model Structure

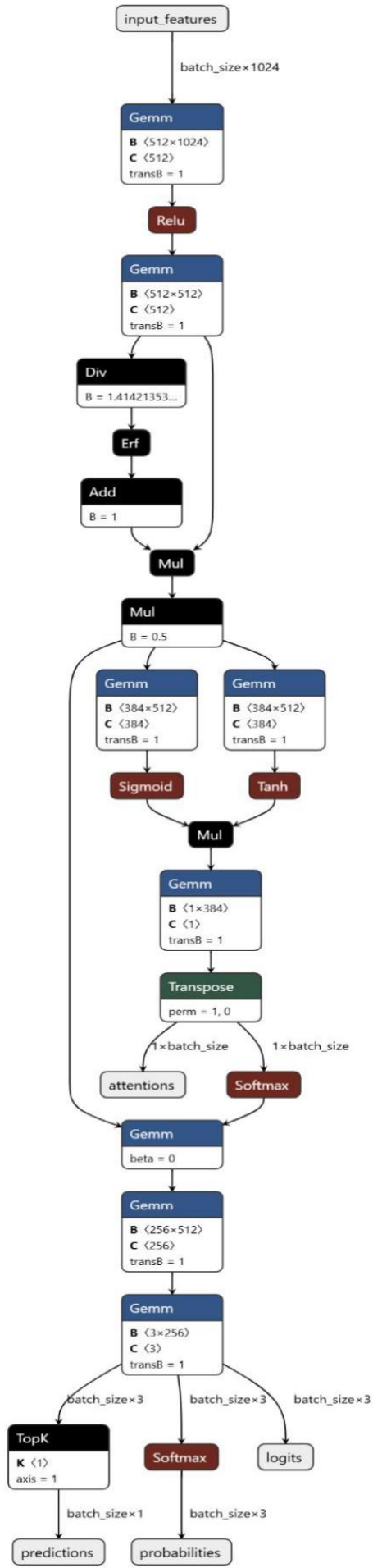

Figure 5. Training workflow of CLAM-SB

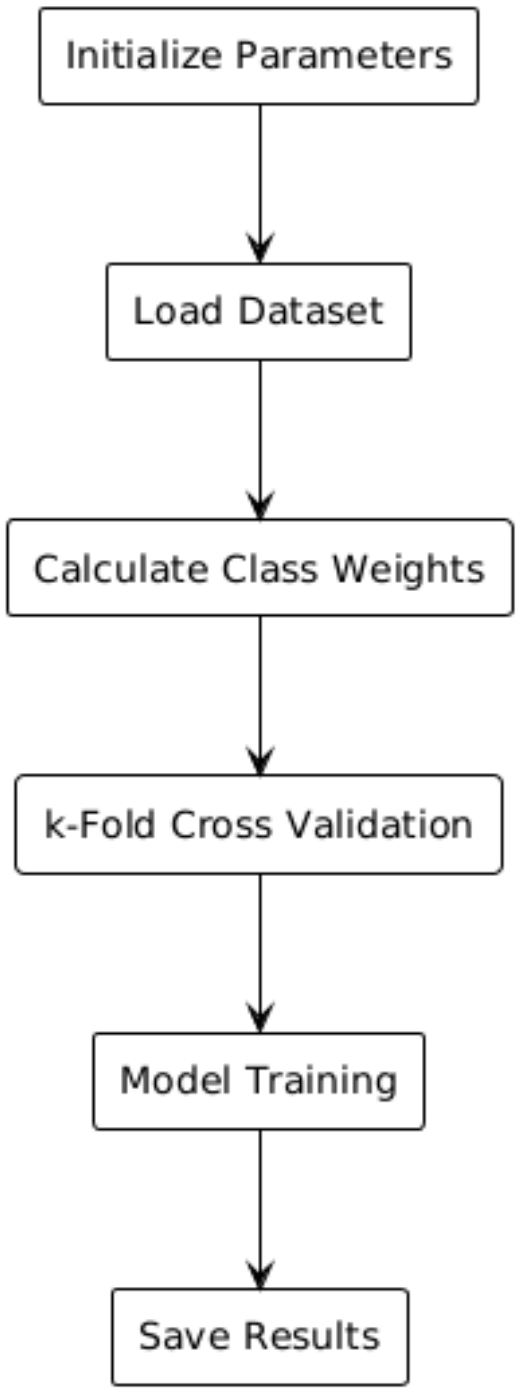

Figure 6. ABMIL model structure

Figure 7. ConvNeXt-MIL-XGBoost structure

Figure 8. Accuracy curve of the optimal CLAM model on the training set

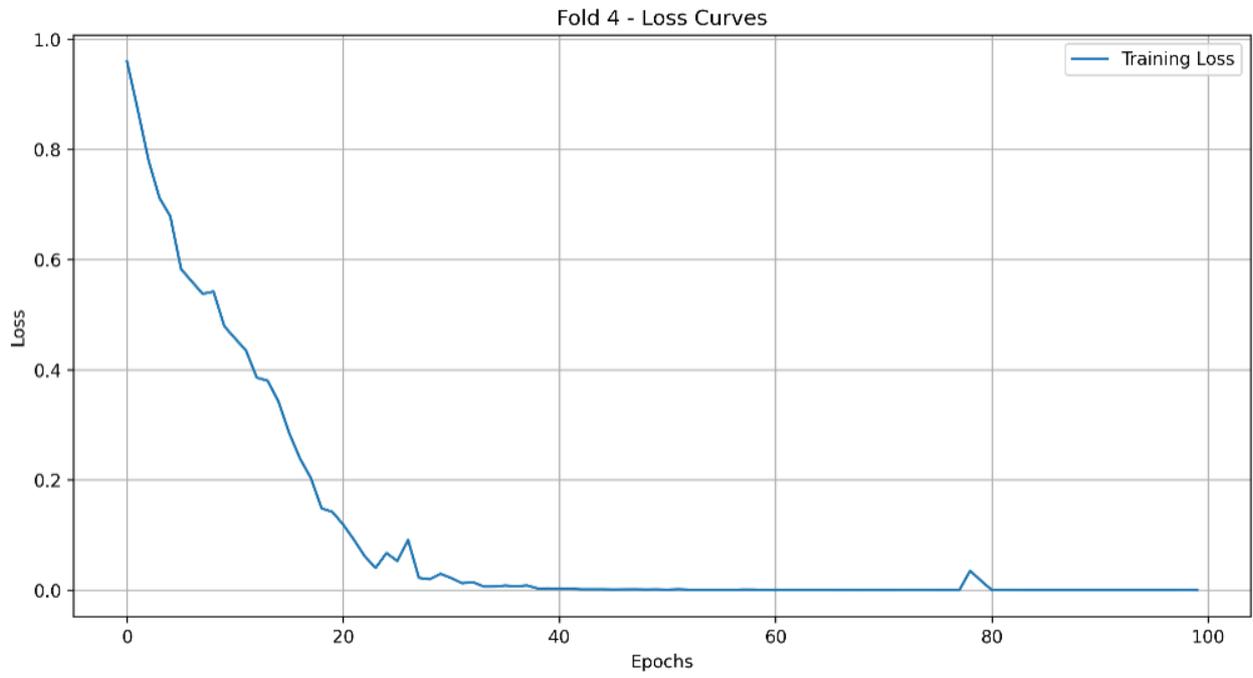

Figure 9. Loss curve of the optimal CLAM model on the training set

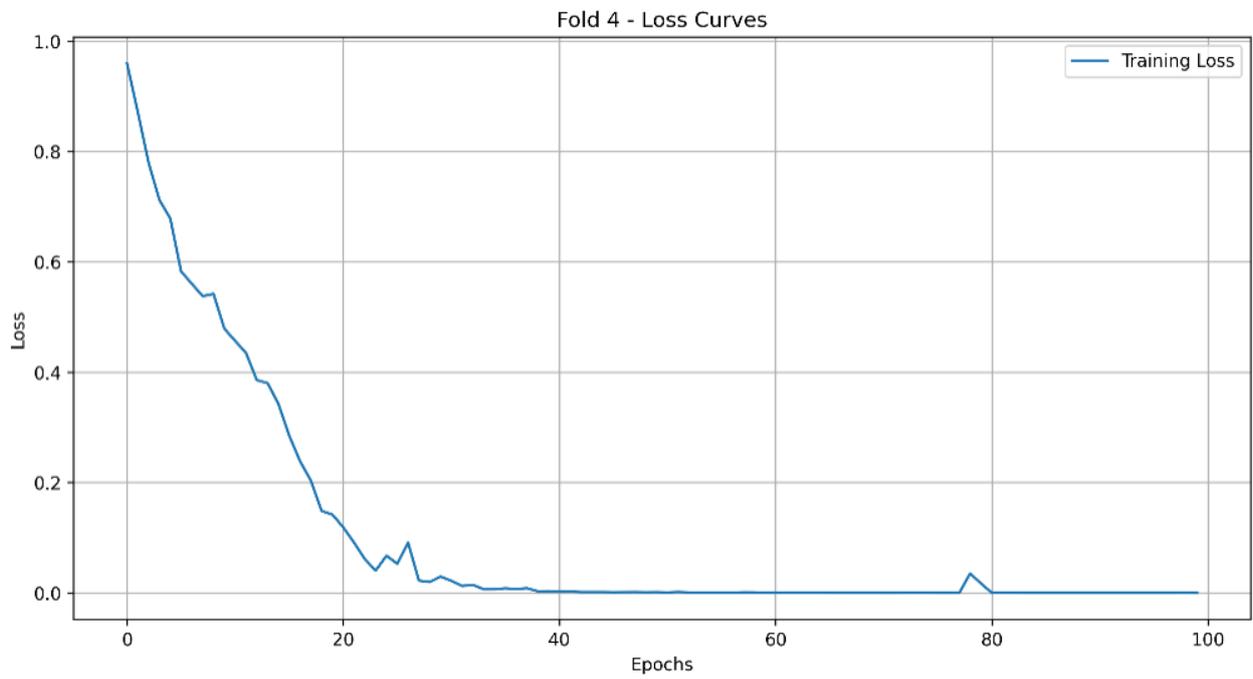

Figure 10. Accuracy-loss curve of the optimal ABMIL on the training set

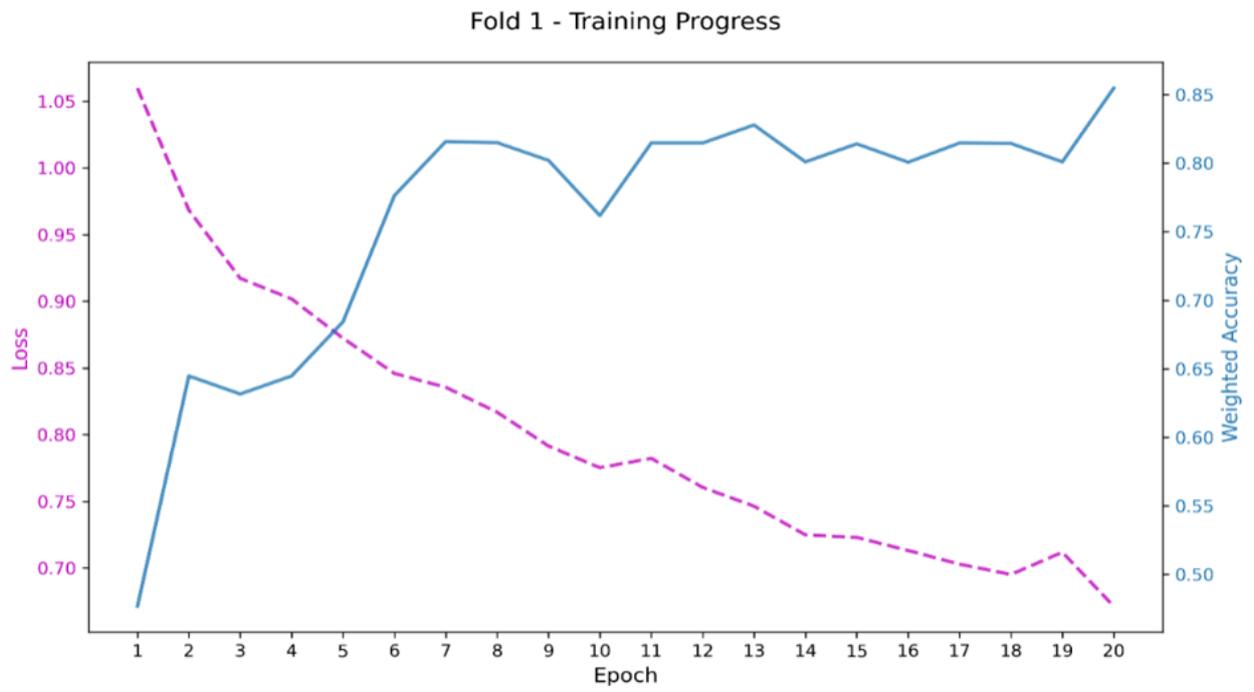

Figure 11. Attention Heatmap of the CLAM Model

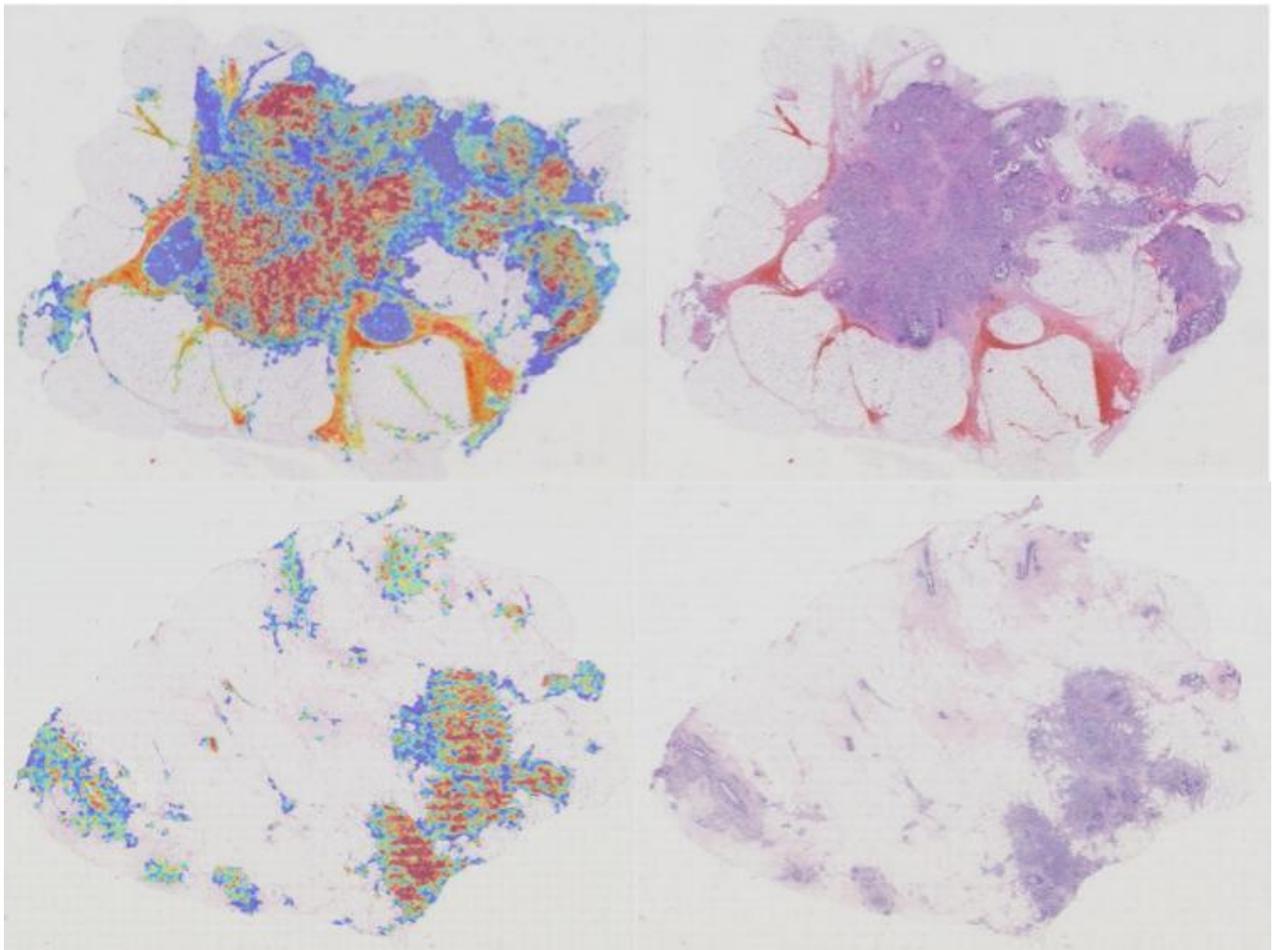

Figure 12. Detailed View of the CLAM Model's Attention Heatmap (Zoomed In)

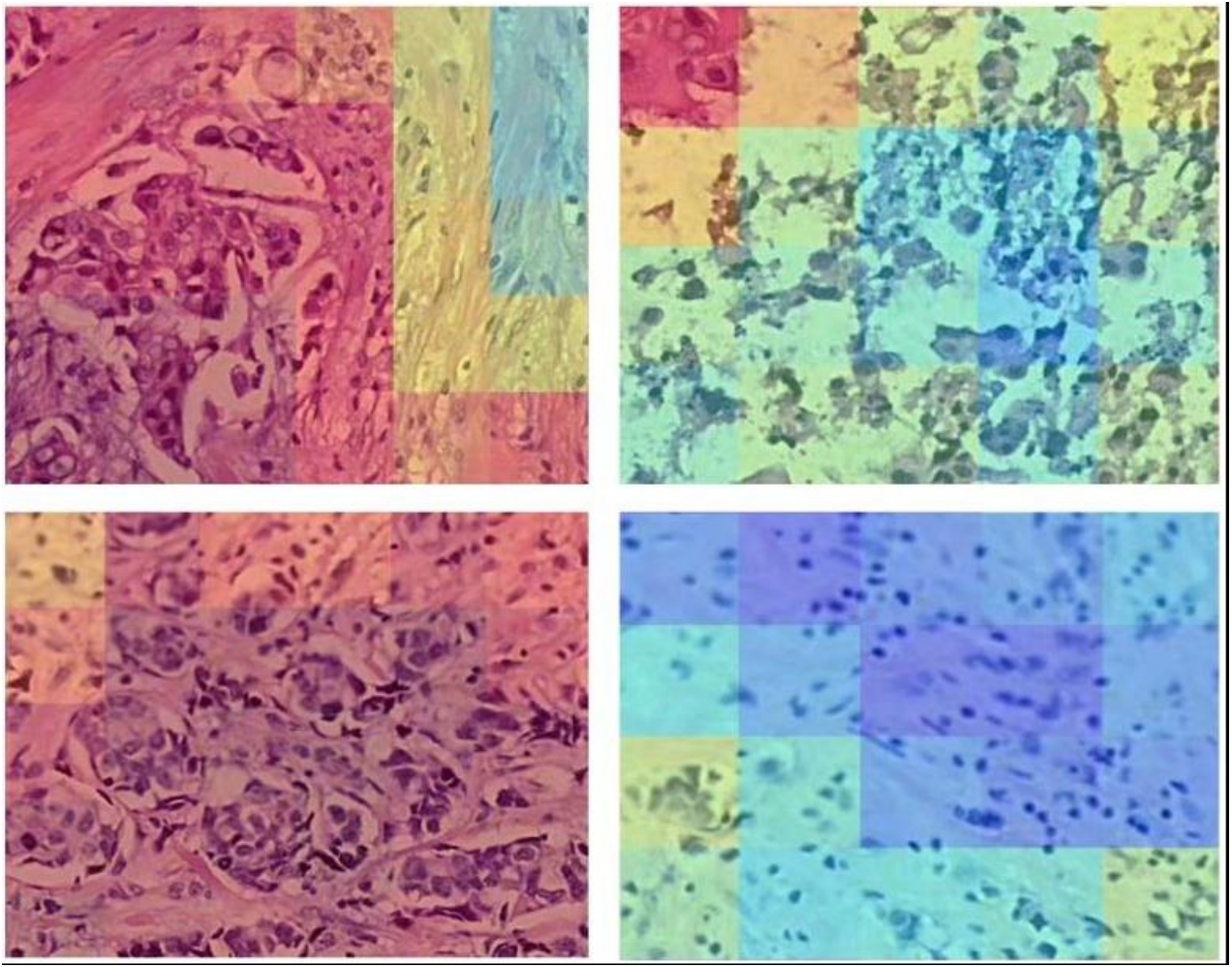

Figure 13. Attention Heatmap of the ABMIL Model

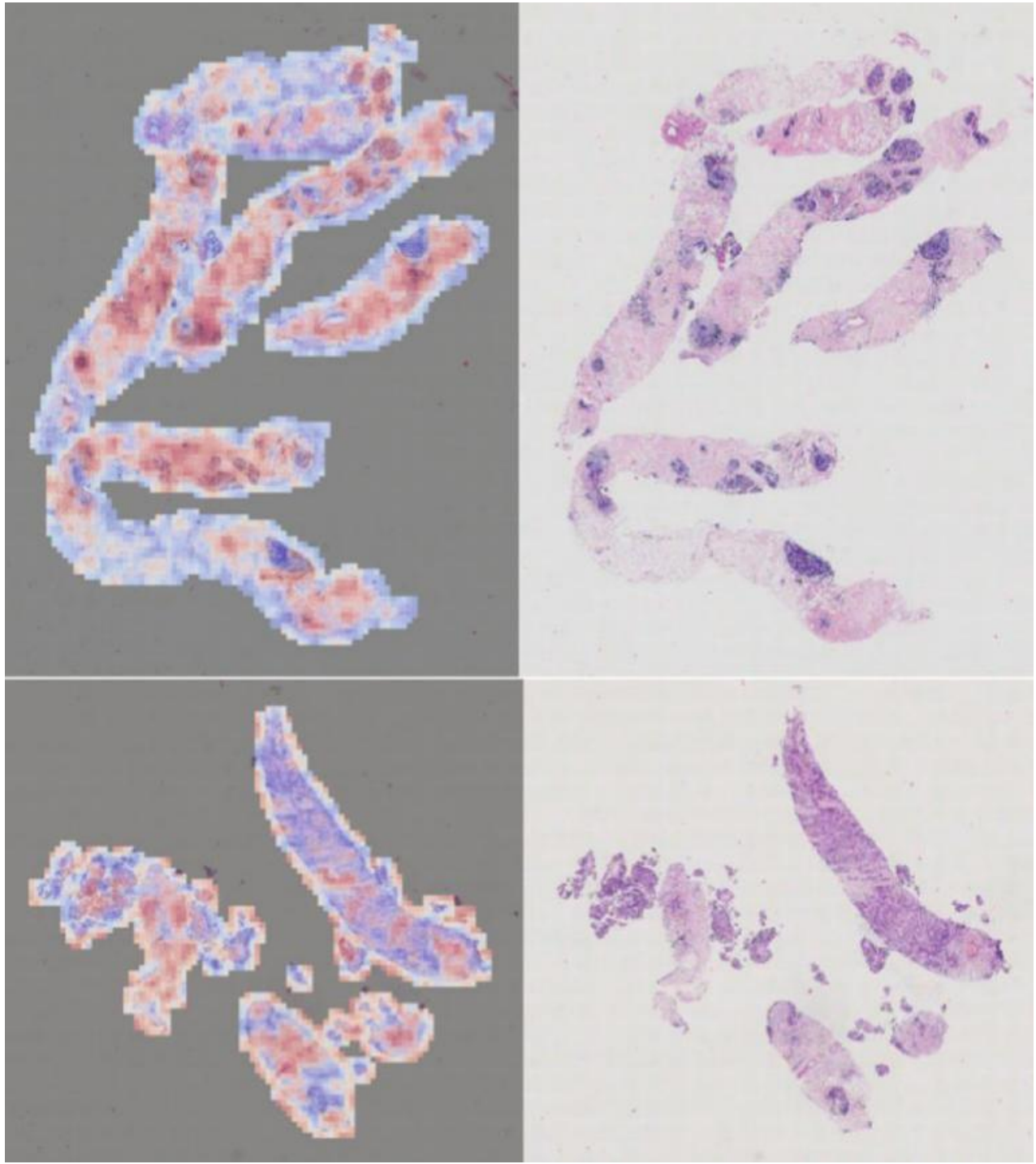

Figure 14. Detailed View of the ABMIL Model's Attention Heatmap (Zoomed In)

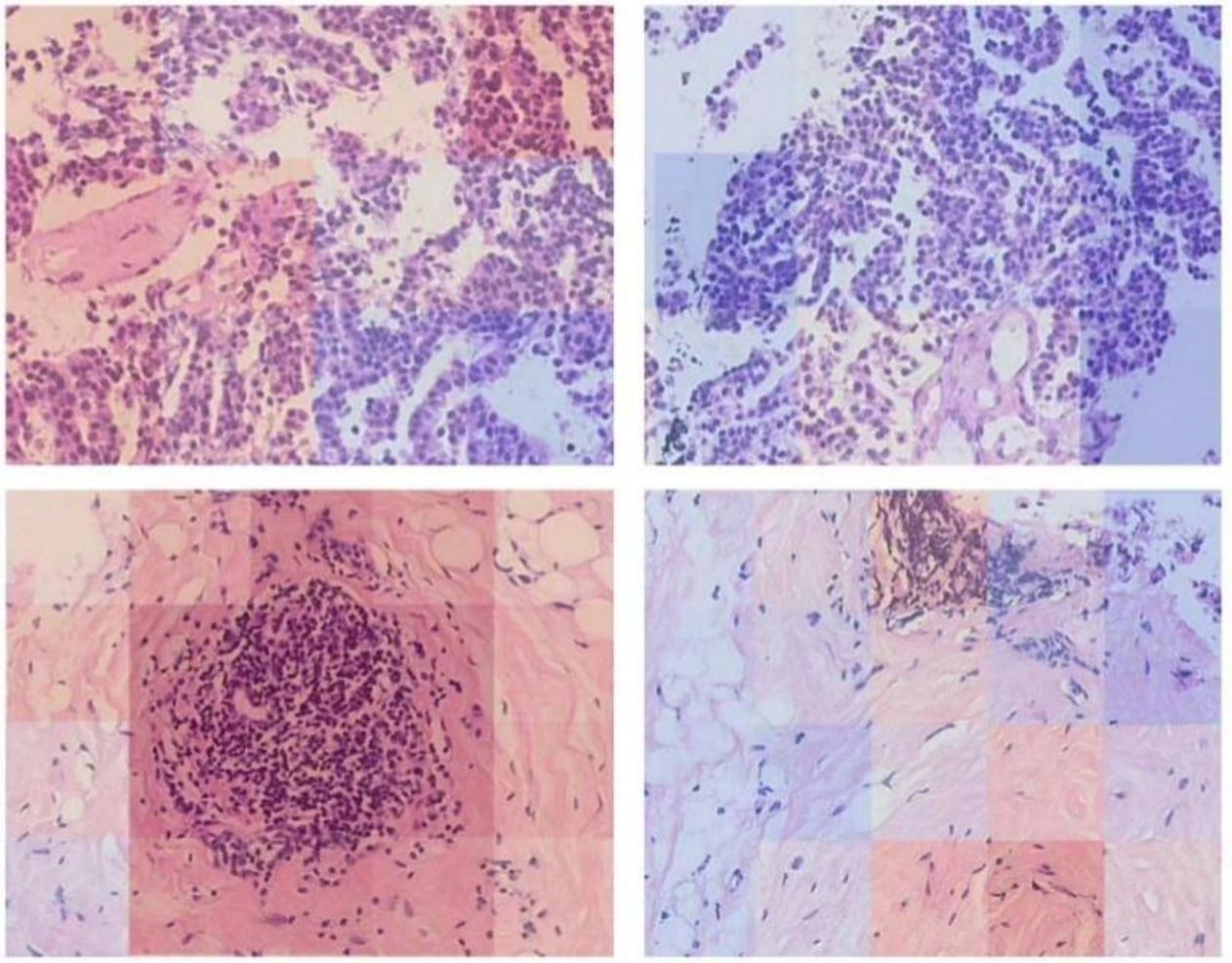

Figure 15. Overall model performance in terms of accuracy and macro F1-score across training, validation, and test datasets.

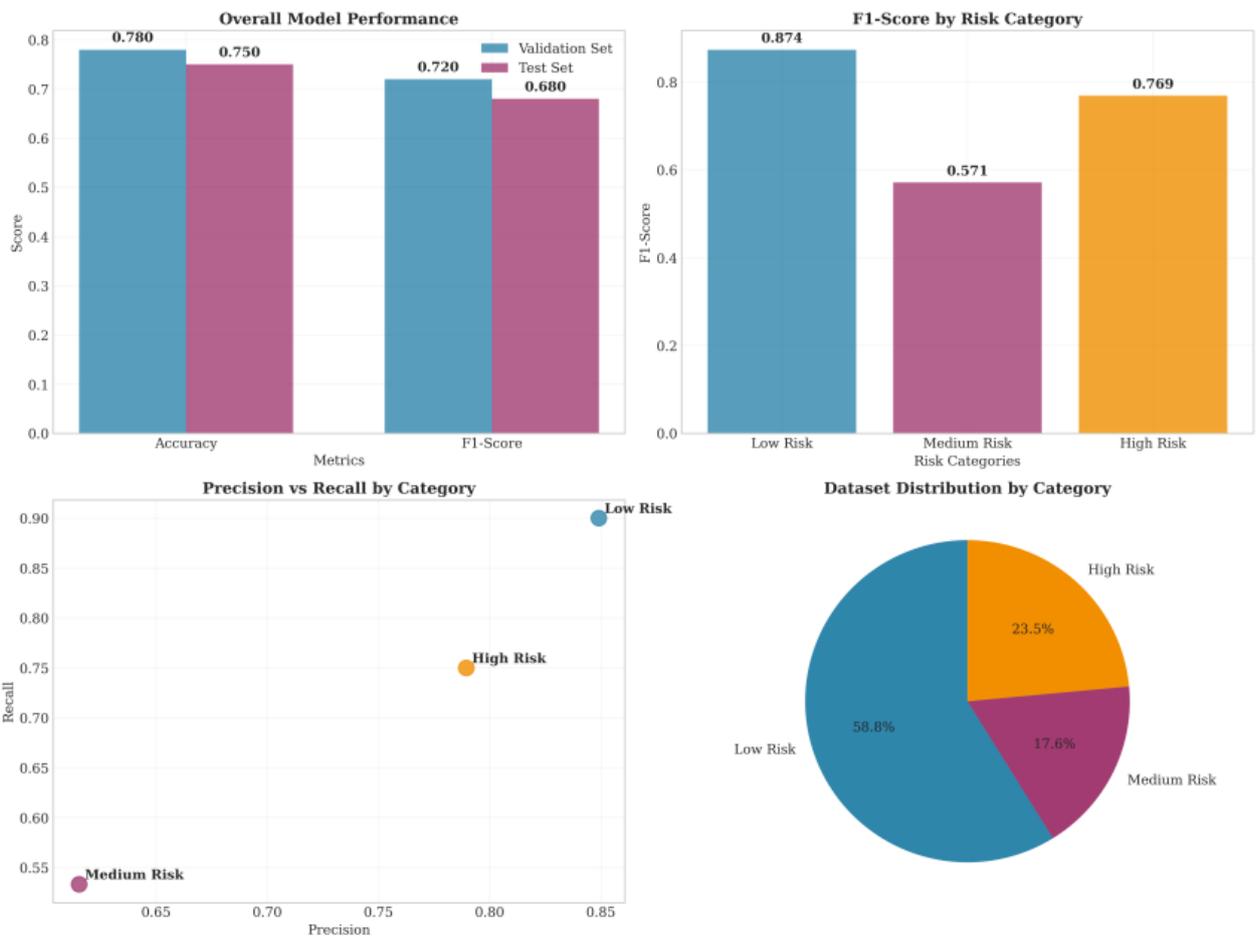

Figure 16. Confusion matrix showing prediction accuracy across the three risk categories in the test set.

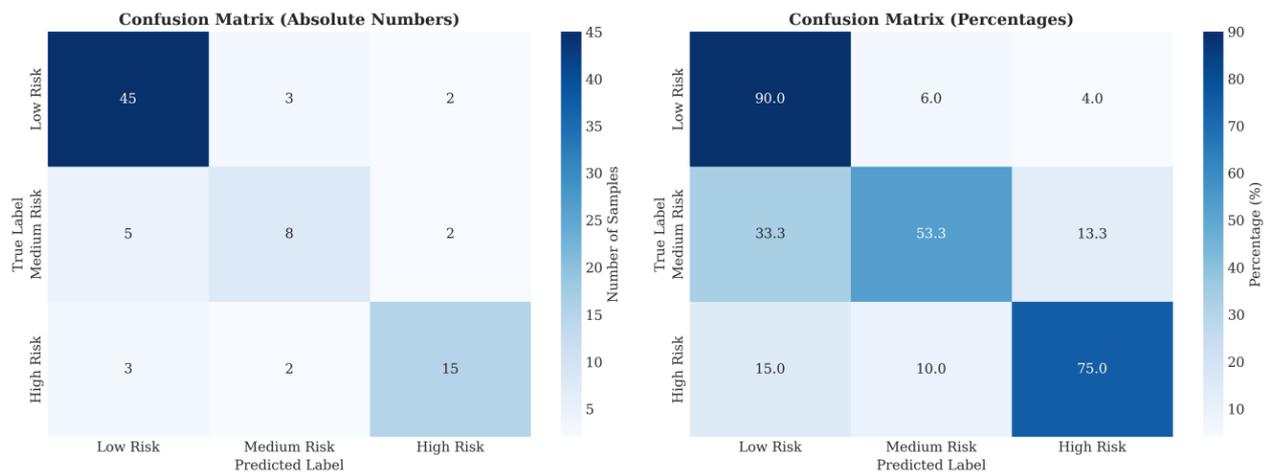

Figure 17. Top feature importance scores derived from the XGBoost classifier trained on the enhanced 23-dimensional feature set.

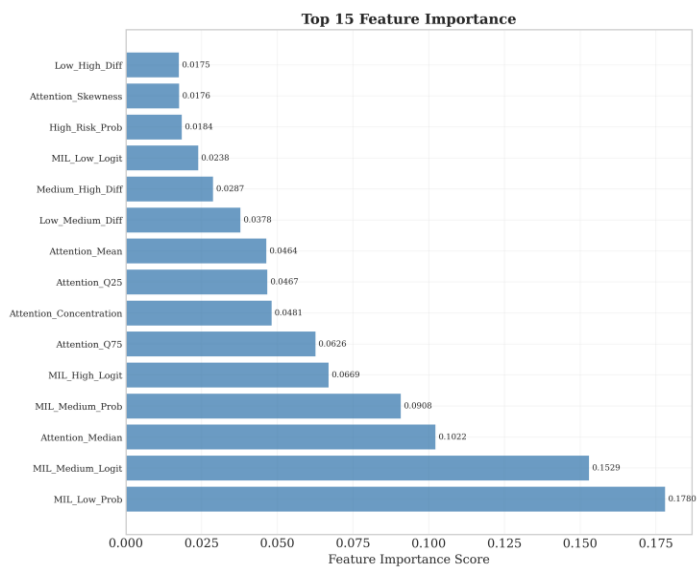

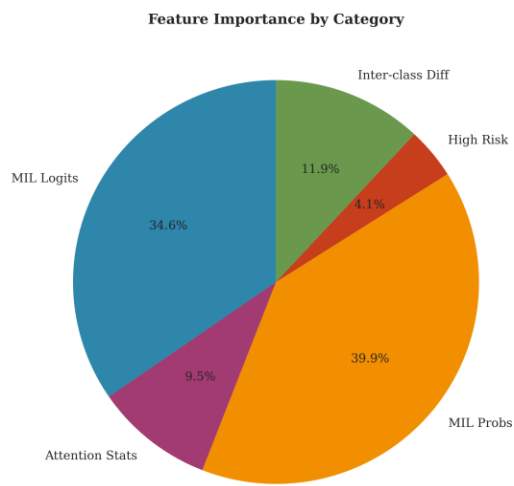

Table 1. Criteria for estimating recurrence risk and treatment plan based on 21-gene score

| Patient Group | Recurrence Score (RS) | Treatment Recommendation |
|---|---|---|
| 21-Gene Assay (Oncotype DX) For Postmenopausal Patients with pN0 or pN1 (1-3 positive nodes) | < 26 | For patients with T1b/c-2, pN0, HR+, HER2- cancer, the prospective TAILORx trial showed no benefit from adding chemotherapy to endocrine therapy for RS 11-25. The RxPONDER trial also showed no chemotherapy benefit for postmenopausal patients with pN1 disease and RS < 26. |
| | ≥ 26 | For patients with pT1-3, HR+, HER2- disease (with pN0 or pN1), adding chemotherapy to the endocrine therapy regimen is recommended. |
| 21-Gene Assay (Oncotype DX) For Premenopausal Patients with pN0 | ≤ 15 | The prospective TAILORx trial showed no benefit from adding chemotherapy to endocrine therapy for patients with T1b/c-2, pN0, HR+, HER2- cancer. |
| | 16 - 25 | A small benefit from adding chemotherapy cannot be excluded, but it is unclear if this is due to chemotherapy-induced ovarian suppression. Options include chemotherapy followed by endocrine therapy, or ovarian function suppression (OFS) combined with tamoxifen or an aromatase inhibitor (AI). |
| | ⩾ 26 | For HR+, HER2-, and pN0 premenopausal patients, adding chemotherapy to endocrine therapy is recommended. |
| 21-Gene Assay (Oncotype DX) For Premenopausal Patients with pN1 (1-3 positive nodes) | < 26 | Compared to endocrine therapy alone, adding chemotherapy provides a small reduction in distant recurrence, though this benefit may be due to chemotherapy-induced ovarian suppression. For this group, options include adding chemotherapy, or using OFS combined with tamoxifen. |
| | ⩾ 26 | For HR+, HER2- premenopausal patients with pT1-3 and pN1 disease, adding chemotherapy to endocrine therapy is recommended. |

Table 2. Parameter settings for training the CLAM model

| Hyperparameter | Value | Description | Rationale |
|---|---|---|---|
| `--lr` | 3e-5 | Small learning rate for fine-tuning | Prevents catastrophic forgetting of pre-trained features, ideal for transfer learning scenarios. |
| `--reg` | 1e-4 | L2 Regularization coefficient | Controls overfitting and balances model complexity with performance. |
| `--drop_out` | 0.4 | Dropout rate (40% of neurons) | Enhances model generalization and compensates for the small dataset size. |

| Hyperparameter | Value | Description | Rationale |
|---|---|---|---|
| `--max_epochs` | 100 | Maximum number of training epochs | Provides sufficient iterations for convergence, used with early stopping to prevent overfitting. |
| `--bag_loss` | Focal | Loss function for bag-level prediction | Automatically adjusts weights for hard examples to address the severe class imbalance. |
| `--bag_weight` | 0.5 | Weighting factor for instance- vs. bag-level loss | Balances the learning objective between local patch-level details and the global slide-level prediction. |
| `-B` | 8 | Number of attention heads | Allows the model to capture diverse feature patterns from different regions simultaneously. |
| `--embed_dim` | 1024 | Dimension of feature embeddings | Ensures sufficient capacity to retain rich feature information from the pre-trained encoder. |
| `--warmup_epochs` | 5 | Linear learning rate warmup epochs | Stabilizes gradients during the initial phase of training to prevent optimization instability. |

Table 3. Parameter Settings for Training ABMIL Model

| Hyperparameter | Value | Description | Rationale |
|---|---|---|---|
| `--lr` | 4e-4 | Learning rate for fine-tuning | Prevents catastrophic forgetting of pre-trained features, suitable for transfer learning scenarios. |
| `--max_epochs` | 20 | Maximum number of training epochs | Provides sufficient iterations for convergence, used with early stopping to prevent overfitting. |
| `--B` | 8 | Number of attention heads | Allows the model to capture diverse feature patterns from different regions simultaneously. |
| `--head_dim` | 512 | Dimension of each attention head | Defines the representational capacity of each attention head. |
| `--input_feature_dim` | 768 | Dimension of input features | Aligned with the output dimension of the upstream feature extractor (CONCH). |

Table 4. AUC and accuracy of the fine-tuned CLAM model on training set

| Fold | Validation AUC | Validation Accuracy |
|---|---|---|
| 0 | 0.774 | 0.714 |
| 1 | 0.939 | 0.810 |
| 2 | 0.679 | 0.667 |
| 3 | 0.929 | 0.810 |
| 4 | 0.952 | 0.810 |
| **Mean** | **0.836** | **0.762** |

Table 5. AUC and accuracy of ABMIL on the test set

| Fold | Validation AUC | Validation Accuracy |
|---|---|---|
| 0 | 0.755 | 0.727 |
| 1 | 0.799 | 0.773 |
| 2 | 0.720 | 0.727 |
| 3 | 0.728 | 0.636 |
| 4 | 0.832 | 0.682 |
| **Mean** | **0.767** | **0.709** |